\documentclass{article}

\usepackage{microtype}
\usepackage{graphicx}
\usepackage{subcaption}
\usepackage{booktabs} 

\usepackage{hyperref}

\usepackage[preprint]{icml2026}

\usepackage{amsmath}
\usepackage{amssymb}
\usepackage{mathtools}
\usepackage{amsthm}

\usepackage{graphicx}
\usepackage{amsmath}
\usepackage{multirow}
\usepackage{svg}
\usepackage{subcaption}
\usepackage{wrapfig}

\usepackage[capitalize,noabbrev]{cleveref}

\theoremstyle{plain}
\newtheorem{theorem}{Theorem}[section]

\theoremstyle{definition}

\theoremstyle{remark}

\usepackage[textsize=tiny]{todonotes}

\newcommand{\method}{\textsc{DotResize}}

\icmltitlerunning{\method: Reducing LLM Width via Discrete OT-based Neuron Merging}

\begin{document}

\twocolumn[
  \icmltitle{\method: Reducing LLM Width via \\Discrete Optimal Transport-based Neuron Merging}

  \icmlsetsymbol{equal}{*}

  \begin{icmlauthorlist}
    \icmlauthor{Neha Verma}{clsp}
    \icmlauthor{Kenton Murray}{clsp,coe}
    \icmlauthor{Kevin Duh}{clsp,coe}
  \end{icmlauthorlist}

  \icmlaffiliation{clsp}{Center for Language and Speech Processing, Johns Hopkins University, Baltimore, MD, USA\\}
  \icmlaffiliation{coe}{Human Language Technology Center of Excellence, 
Johns Hopkins University, Baltimore, MD, USA}

  \icmlcorrespondingauthor{Neha Verma}{nverma7@jhu.edu}

  \icmlkeywords{Optimal Transport, LLM Pruning, Merging}

  \vskip 0.3in
]

\printAffiliationsAndNotice{}  %

\begin{abstract}
Structured pruning methods designed for Large Language Models (LLMs) generally focus on identifying and removing the least important components to optimize model size.
However, in this work, we question this prevalent approach by instead exploring how to recombine information from structures designated for pruning back into the reduced model. %
We specifically focus on neuron width reduction, and frame this problem as a \textbf{D}iscrete \textbf{O}ptimal \textbf{T}ransport problem, and propose \method, a novel Transformer compression method that uses optimal transport theory to transform and compress model width. %
To ensure applicability within the Transformer architecture, we motivate and incorporate necessary entropic regularization and matrix factorization techniques into the transportation maps produced by our method. %
Unlike pruning-based approaches which discard neurons based on importance measures, \method~ re-projects the entire neuron width, allowing the retention and redistribution of useful signal across the reduced layer. %
Empirical results show that compared to simple or state-of-the-art neuron width-pruning techniques, \method~ serves as a useful add-on to pruning, while achieving measurable reductions in real-world computational cost. %
\end{abstract}

\section{Introduction}

High-quality compression techniques for LLMs have become increasingly important due to the ubiquity of pre-training and success of model and data scaling. %
Compression can reduce model size and latency, which helps facilitate reduced inference costs, on-device capabilities, and improved access to models by both users and researchers. %

Structured pruning is a popular approach to compression because it directly reduces the on-disk size of LLMs and accelerates inference, without requiring specialized hardware or libraries, often needed for unstructured pruning.
By targeting and removing entire structures, such as attention heads, blocks from weights, neurons, or entire layers, structured pruning delivers direct inference speed-ups and memory savings. 
One variety of structured pruning reduces the width of weights across model layers, resulting in compressed models with more computational throughput \citep{ma2023llm, ashkboos2024slicegpt}. 
In general, these width pruning techniques aim to identify important neurons, and discard neurons which are deemed less important. 

Beyond weight removal, a complementary line of research focuses on applying invariant transformations to model parameters. 
These methods involve designing functions $\pi$ that modify model parameters $\theta$ such that the model preserves $f(x|\theta) = f(x|\pi(\theta))$, to prepare the model for downstream tasks. 
Prior work has designed invariant transformations to enable many model merging methods \citep{singh2020model,ainsworthgit}, and has recently been explored for single-model compression applications \cite{theustowards}. 
These single-model applications tend to design functions $\pi^*$ such that the model function is approximated due to the compression objective, i.e. $f(x|\theta) \approx f(x|\pi^*(\theta))$. 
Relevant examples include QuaRot, where rotational matrices are applied to Transformer weights to reduce outliers for improved low bit quantization \citep{ashkboos2024quarot}, and SliceGPT, where rotational PCA matrices are again applied to prepare weights for low-eigenvalue neuron pruning \citep{ashkboos2024slicegpt}.

\begin{figure*}[t]
    \centering
    \includegraphics[width=0.8\linewidth]{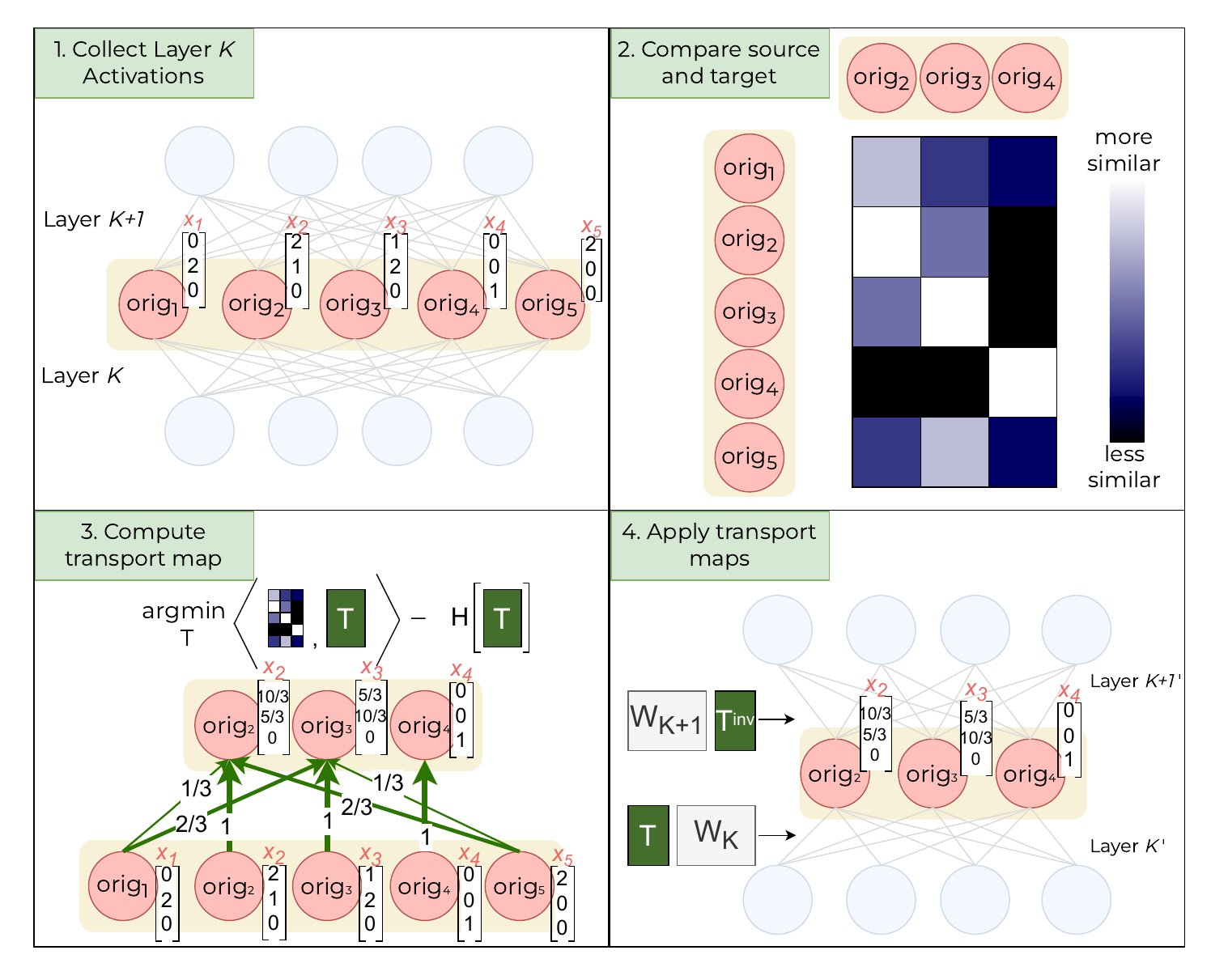}
    \caption{A depiction of our neuron width merging strategy. In panel 1, we demonstrate computing activations from layer $K$ in preparation for panel 2, where we select a subset of 3 neurons from this layer, and compute pairwise similarities between the activations of the 5 original neurons, and the activations of the subset. In panel 3, we compute the optimal transport map, depicted in green, by optimizing the map according to the similarities and entropic regularization. Finally, we demonstrate replacing layer $K$ with the subset of neurons, after transforming its weights with $T$ and layer $K+1$'s weights with $T^{\text{inv}}$, resulting in new activations. }
    \label{fig:overview}
    \vspace{-0.2cm}
\end{figure*}

In this work, we move beyond the binary choice of keeping or discarding neurons inherent to pruning by proposing an merging-based augmentation to the common neuron pruning paradigm \cite{ma2023llm, ashkboos2024slicegpt}. We introduce  a method that merges signal from neurons to be pruned into the remaining width using optimal transport. This approach represents a first step toward unifying model merging and structured pruning, aiming to reincorporate the ``unaccounted'' signal that is typically lost during compression. 
While a prior transport-based approach has been proposed and applied to ResNet-based image models \citet{theustowards}, we identify and propose solutions to overcome the key architectural hurdles required to apply these techniques to LLMs. 
Our method does not require recovery fine-tuning, offering a promising new direction for efficient, zero-shot model compression.

We find that when applied atop of gradient-free LLM neuron pruning methods, \method~frequently recovers additional performance by re-incorporating signal from neurons that would have otherwise be pruned.
Our method is summarized in Figure \ref{fig:overview}.\footnote{Full code including an implementation of our method and experiments will be released upon publication. }

Our contributions can be summarized as follows.
\begin{enumerate}
\item We propose a novel, training-free neuron width reduction method in LLMs that uses entropy-regularized discrete optimal transport to recombine signal from neurons that would otherwise be pruned from neuron pruning methods.
\item We re-examine a specific invariance result found in \citet{ashkboos2024slicegpt} that enables the application of orthogonal maps throughout LLM weights. We extend this set of invariant maps from orthogonal to any invertible map,  via the application of QR-decomposition to respect the orthogonal-only invariance of RMSNorm. 
\item We demonstrate the effectiveness of \method~ by comparing our method directly to similar and performant pruning alternatives on several LLMs. We show that across language modeling and zero-shot LLM tasks, our method can frequently regain some of the performance loss caused by neuron-level pruning in directly comparable methods, supporting the core motivation of our work. 

\end{enumerate}

\section{Related Work}

\paragraph{Neuron-level pruning} 
Given the practical limitations of unstructured weight pruning \citep{lecun1989optimal}, including the lack of easily realizable real-world memory and latency savings, recent work on weight pruning has examined converting unstructured sparsity patterns into semi-structured sparsity patterns \citep{frantar2023sparsegpt, sunsimple}. 
However, this conversion process frequently leads to additional performance loss and only provides speed-ups on specific GPU architectures.  
Alternatively, structured pruning techniques that remove structured portions of models, like attention heads, layers, individual neurons, and blocked weight regions, provide a hardware agnostic approach to achieving both real-world disk and latency savings. 
In this work, we focus on reducing individual neurons, which has been explored in prior work via neuron width pruning. 
\citet{ma2023llm} propose a ``channel pruning'' technique that removes neurons sharing the same index across all LLM layers, effectively reducing the width of the entire model. \citet{ashkboos2024slicegpt} propose a related technique, but use PCA projections to perform the width reduction to prune low eigenvalue associated neurons in the re-projected eigenspace.
\citet{gao2024disp} reduce neuron width in a dimension-independent manner by using different indexing matrices per layer. 
Their method trains a separate GRU hypernetwork to predict neurons to remove. 
We depart from these pruning approaches by instead focusing on merging-based strategies. 

\paragraph{Neuron-level redundancy} 
Current LLMs are highly overparameterized, containing more parameters than necessary to fit their training data. This leads to consequences like the redundant encoding of concepts, which has been repeatedly identified in prior work. 
\citet{srinivas2015data} propose a method to wire similar neurons together within simple networks based on their weight set similarities. 
\citet{dalvi-etal-2020-analyzing} analyze redundancy between individual neurons and layers and propose a clustering-based approach to pruning neurons. 
In a model merging method, \citet{stoicazipit} tackle redundancy by not only aligning similar neurons \textit{across} models to merge, but also align similar neurons \textit{within} a model for merging. 
\citet{nanda2023diffused} characterize redundancy in pre-trained models as ``diffuse,'' where multiple random subsets of neurons in a model layer can approximate the performance of the whole layer. In our work, we are inspired by this diffusely redundant nature of neurons, and design our technique to target potentially redundant groups of neurons within a layer. \citet{theustowards} use optimal transport to merge groups of neurons specifically in ResNets, where only certain residual groups undergo optimal-transport based compression. We extend ideas in this approach to fully residually-connected LLMs, and operate at a layer level rather than group-level.

\section{\method}
\label{sec:method}

Our \method~width reduction method draws on entropy-regularized discrete optimal transport, where we assign a given layer's neurons to a source distribution and a subset of these neurons as a target distribution, and compute an optimal transport map between source and target based on activation signatures over representative data. 
The result of this approach is a method that reduces overall model width by transforming the signal from the entire neuron width, which is in contrast with pruning methods that only consider a portion of the signal. A demonstration of this width reduction is displayed in Figure \ref{fig:overview}. Reducing model width incidentally reduces activation width too, improving activation memory use.

\subsection{Computing the transport maps}

Discrete optimal transport problems are formulated as transforming a source distribution $\mathbf{a}$ to a target distribution $\mathbf{b}$ subject to a cost matrix that dictates the cost of transforming each source item to each target item, denoted as $C$. The goal of optimal transport is to find a map that minimizes this cost. 

Following \citet{singh2020model}, we define our source and target distributions as uniform, with the source distribution as $\mathbf{a} = \mathbf{1}_{d_{\text{orig}}} / d_{\text{orig}} $, and target distribution as $\mathbf{b} = \mathbf{1}_{d_\text{new}} / d_\text{new}$, where we select $d_{\text{new}} < d_{\text{orig}}$ neurons from the original width. 
We assume that we are equipped with an underlying pruning strategy that selects these neurons. 
We define the cost matrix of transporting mass from distribution $\mathbf{a}$ to distribution $\mathbf{b}$ as the pairwise distances between the activations of the $d_{\text{orig}}$ source neurons, and the activations of the  $d_\text{new}$ selected neurons. 
This means that at each layer we wish to compress, we compute activations across $n$ exemplar tokens and use these activations to help define our ground metric. 
In our work, we use $\ell_1$ norm to compute pairwise distances.
This results in cost matrix $C \in \mathbb{R}^{d_\text{orig} \times d_\text{new}}$, where for activation rows $x_i, x_j \in \mathbb{R}^{n}$ corresponding to source neuron $i$ and target neuron $j$, $C_{ij} = || x_i - x_j ||_1$. 

We find our transportation maps via an entropy-regularized objective \cite{cuturi2013sinkhorn}. 
We use Sinkhorn regularization to encourage smoothing of the transportation map, such that each target neuron can be a combination of multiple source neuron, rather than just a few. 
Prior work has found Sinkhorn regularization to be critical when merging separate Transformers via transport mapping \cite{imfeldtransformer}. 
Additionally, Sinkhorn regularization improves the computational tractability of the OT objective.

The following optimization problem reflects the Sinkhorn objective for finding map $T$ between distributions $\mathbf{a}$ and $\mathbf{b}$ subject to costs $C$. 
\begin{align}
T =& \arg\min_{T} \langle T, C \rangle  - \lambda H(T) \notag \\
&\text{s. t. } T\mathbf{1}_{d_{\text{orig}}} = \mathbf{a}, T^T\mathbf{1}_{d_{\text{new}}} = \mathbf{b}
\label{eq:1}
\end{align}
The larger the parameter $\lambda$, the ``softer'' the transportation map is due to the flattening effect of the entropy term. 

After applying map $T$ to the one or more weight matrices that produced the activations used for alignment, we can apply an inverse transformation to the following weight matrix such that functional invariance is preserved. In practice, we must modify these maps in order to interface with the Transformer architecture, as described in following sections.

\subsection{Maintaining invariance via QR decomposition}
\label{method:qr}

\begin{figure}[t]
\begin{center}
    \includegraphics[width=0.46\textwidth]{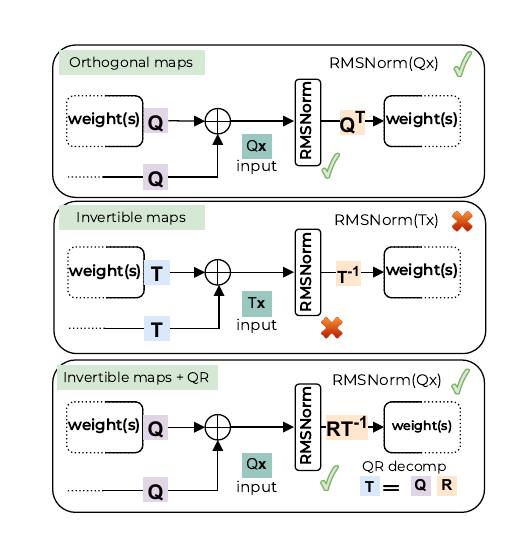}
    \caption{Our QR-decomposition step allows for general invertible matrices to apply at residual junctions as depicted by the figure. The figure depicts a general residual connection as $\bigoplus$ in a pre-norm Transformer layer. While orthogonal transformations are naturally invariant to RMSNorm, general invertible matrices are not unless decomposed via QR decomposition and re-routed as shown. The associativity of matrix multiplication allows us to absorb matrix R into the $T^{-1}$ calculation, allowing the orthogonal multiplicand to not change RMSNorm.}
    \label{fig:qr_decomp}
\end{center}
\end{figure}

In applying the transportation matrices to transform model weights, we aim to approximate the original model's function to achieve compression without losing substantial performance. 
In \citet{ashkboos2024slicegpt}, orthogonal PCA matrices are used to transform model weights given their invariance to the commonly used root mean square layer normalization (RMSNorm) in Transformers \citep{zhang2019root}. 
This invariance allows for transformation matrices to apply to weights before and after a residual connection, while not interfering with the function of RMSNorm in pre-norm architectures.
For orthogonal matrix $Q$ and data vector $\mathbf{x}$, this invariance can be expressed as the following:
\begin{equation}
\text{RMSNorm}(\mathbf{x}Q)Q^T = \text{RMSNorm}(\mathbf{x})
\end{equation}
A depiction of this invariance is found in the first panel of Figure \ref{fig:qr_decomp}. 
This invariance assumes that RMSNorm is unweighted, as weighted versions can be converted to unweighted RMSNorms by pre-multiplying weight matrices following pre-norm by the RMSNorm weights \cite{elhage2023privileged}. For more details on this conversion, refer to \citet{ashkboos2024slicegpt}.

To apply our method, we must relax the constraint of permissible transportation matrices from only orthogonal matrices to any invertible matrix. 
While the use of orthogonal matrices in prior work was important as they do not change vector norms, general invertible matrices may change norms.
To address this limitation of invertible transformations, we use a QR decomposition step in order to decompose invertible matrix $T = QR$ such that the $Q$ matrix is applied to the input vector before  RMSNorm, and the $R$ matrix is applied after the RMSNorm. 
Given decomposition $T = QR$, inverse matrix $T^{-1}$ and data vector $\mathbf{x}$, we have the following relationship:
\begin{align}
\text{RMSNorm}(\mathbf{x}Q)RT^{-1} = \frac{\mathbf{x}Q}{\lVert \mathbf{x}Q\rVert} RT^{-1}& \notag \\ = \frac{\mathbf{x}QRT^{-1}}{\lVert \mathbf{x}\rVert}  = \text{RMSNorm}(\mathbf{x}) &\label{eq:qr_invariance}
\end{align}
A simple explanation of steps in this equality appears in Appendix \ref{app:qr_proof}. 
Given the associativity of matrix multiplication, we are able to QR-decompose our computed transport map and then store just $M = Q$ and $M^{\text{inv}} = RT^{-1}$ as our final transport and untransport maps. An overview of the application of these maps is summarized in Figure \ref{fig:qr_decomp}. 

While in the case of square $T$ this equality holds, we note that our transport maps are rectangular due to the intentional difference in the width of source and target supports. 
Rectangular $T$ is not invertible, meaning that we instead use the pseudoinverse instead in our method.
We show that up to orthogonal projection onto $\text{col}(Q)$ this equality holds in Appendix \ref{app:rectangular_qr_proof}, and in practice, this QR modification is vital for stability; without it, our method fails to converge as the preservation of layer normalization is lost.
Despite our specific application of these invariant maps in our setting, we believe this technique to be broadly useful in other methods that transform models in-place using invariant or approximately invariant maps. 

\subsection{Applying maps}
\label{sec:applying_maps}
With this QR modification, we are able to apply our transformations at two key locations for each Transformer layer \cite{vaswani2017attention}. We label our transformations per layer as $\{M_A, M_A^{\text{inv}}\}$ and $\{M_F, M_F^{\text{inv}}\}$. 

We label each of the linear weight matrices in modern LLMs as the following:
\begin{itemize}
    \item $W_Q ,W_K, W_V$ are the $d_{\text{orig}} \times d_{\text{orig}}$ weight matrices for queries, keys, and values in multi-headed attention (or grouped-query attention \cite{ainslie2023gqa}, where dimensions are $d_{\text{orig}} \times (d_{\text{orig}} / n_\text{groups})$) 
    \item $W_O$ is the output projection of attention, size  $d_{\text{orig}} \times d_{\text{orig}}$. Attention is abbreviated $\text{MHA}(Q, K, V)$
    \item $W_{\text{up}}$ and $W_{\text{gate}}$ are the $d_{\text{orig}} \times d_{\text{ff}}$ up-projection matrices in feed-forward layers with GLU-based activations, which we denote as $\sigma$ for the nonlinearity and $\odot$ for gating \cite{shazeer2020glu}. 
    \item $W_{\text{down}}$ is the $ d_{\text{ff}} \times d_{\text{orig}} $ down-projection matrix for feed-forward layers.  
\end{itemize}
For each layer, we apply $M_A$ and $M_A^{\text{inv}}$ such that the output of attention is multiplied by $M_A$, but is then passed firstly through $M_A^{\text{inv}}$ before being passed through RMSNorm and multiplied by $W_{\text{up}}$ and $W_{\text{gate}}$. We apply $M_F$ and $M_F^{\text{inv}}$ such that the output of the feed-forward layer is multiplied by $M_F$, and then passed firstly through RMSNorm and $M_F^{\text{inv}}$ before being multiplied by $W_Q ,W_K$ and $W_V$. 
In summary, assuming input $x_\text{in}^{\text{attn}}$ is pre-multiplied by $M_F$ to give $M_Fx_\text{in}^{\text{attn}} = x_\text{in}^{\text{attn}\prime}$ from a prior layer or from the input embedding layer, we have:
\begin{align}
x_\text{out}^{\text{attn}} = W_O\text{MHA}(&W_Q M_F^{\text{inv}}x_\text{in}^{\text{attn}\prime}, \notag
\\&W_K M_F^{\text{inv}}x_\text{in}^{\text{attn}\prime}, \notag \\&W_VM_F^{\text{inv}}x_\text{in}^{\text{attn}\prime}) + M_F^{\text{inv}}x_\text{in}^{\text{attn}\prime}
\end{align}
This reverses the effect of pre-multiplying $x_\text{in}^{\text{attn}}$ by $M_\text{F}$ from the prior layer. To apply the $M_A$ compression map, we apply it to $x_\text{out}^{\text{attn}}$, resulting in:
\begin{align}
M_Ax_\text{out}^{\text{attn}} = M_AW_O\text{MHA}(&W_Q M_F^{\text{inv}}x_\text{in}^{\text{attn}\prime}, \notag
\\&W_K M_F^{\text{inv}}x_\text{in}^{\text{attn}\prime}, \notag \\&W_VM_F^{\text{inv}}x_\text{in}^{\text{attn}\prime}) + M_AM_F^{\text{inv}}x_\text{in}^{\text{attn}\prime}
\end{align}
Now, to reverse the effect of $M_A$ on $x_\text{out}^{\text{attn}}$, which we will denote as $x_\text{in}^{\text{ff}\prime} = M_Ax_\text{out}^{\text{attn}}$, we multiply our ff input by $M_A^{\text{inv}}$:
\begin{align}
x_{\text{out}}^{\text{ff}} = W_{\text{down}} \sigma &(W_{\text{up}} M_A^{\text{inv}} x_{\text{in}}^{\text{ff}\prime} \notag \\
&\odot W_{\text{gate}} M_A^{\text{inv}} x_{\text{in}}^{\text{ff}\prime})  + M_A^{\text{inv}} x_{\text{in}}^{\text{ff}\prime}
\end{align}
And finally, to apply the $M_F$ compression map, we have
\begin{align}
M_Fx_\text{out}^{\text{ff}}  = M_FW_\text{down}\sigma&(W_\text{up}M_A^{\text{inv}}x_\text{in}^{\text{ff}\prime} \notag \\
& \odot W_\text{gate}M_A^{\text{inv}}x_\text{in}^{\text{ff}\prime}) + M_FM_A^{\text{inv}}x_\text{in}^{\text{ff}\prime}
\end{align}

All $M$ matrices can be ``absorbed'' into neighboring weight matrices $W$ due to the associativity of matrix multiplication. This application of maps mirrors several prior methods using a similar project/unproject scheme for Transformer model merging and compression \cite{imfeldtransformer, ashkboos2024slicegpt, verma2024merging}.

\section{Experimental Settings}
\label{sec:experimental_settings}

\subsection{Models}

We focus our evaluation on three different families of LLMs: Llama 3.1 at 8B and 70B parameters \citep{grattafiori2024llama}, Mistral at 7B (v0.3) and 12B (NeMo) parameters \citep{yang2024qwen2}, and Phi-4 at 14B parameters \citep{abdin2024phi}. 
These models all reflect large Transformer models pre-trained on primarily English text data. 
We obtain all models using their HuggingFace \texttt{transformers} library implementations \citep{wolf-etal-2020-transformers}. 
All models with the exception of Llama-3.1-70B can be compressed and evaluated on a single 32GB V100 GPU, whereas we use 8 V100s for Llama-3.1-70B experiments. 

\subsection{Evaluation}

For language modeling capabilities, we evaluate our models by computing perplexity on the English Wikitext-2 dataset \citep{merity2017pointer}. 
For zero-shot general capabilities, we test the models on 5 evaluation sets sourced from \texttt{lm-evaluation-harness} \citep{eval-harness}: ARC-Challenge and ARC-Easy \citep{clark2018think}, HellaSwag \citep{zellers2019hellaswag}, PIQA \citep{bisk2020piqa}, and Winogrande \cite{sakaguchi2021winogrande}. 
These five tasks reflect diverse and commonly used English tasks for evaluating language models. 
Following \citet{ashkboos2024slicegpt}, in all language modeling experiments, we use a small subset (128 samples) of the Wikitext-2 training data to compute our transportation maps. Samples are length 2048. For all zero-shot experiments, we use the same number of tokens, but from the Stanford Alpaca dataset for instruction following \cite{alpaca}. 

\begin{table*}[t]
\caption{Perplexity performance of models on Wikitext-2 under different width reduction strategies and at different levels of width reduction. Perplexity results that reflect $>$ 0.1 PPL reduction are bolded. In comparing Magnitude Pruning to \method~ without pre-projection, we see a stark improvement in perplexity, indicating that using optimal transport to reduce neuron width can be superior to solely pruning neurons. \\}
\centering
\begin{tabular}{llrrrrr}
\toprule\toprule
\multirow{2}{*}{Width Reduction} & \multirow{2}{*}{Method} & \multicolumn{2}{c}{Llama-3.1} & \multicolumn{2}{c}{Mistral} & Phi-4 \\
\cmidrule(lr){3-4} \cmidrule(lr){5-6} \cmidrule(lr){7-7}
 &  & 8B & 70B & 7B  & 12B  & 14B \\
 \midrule
 0\% & - & 6.24 & 2.81 & 5.32 & 5.74 & 6.46 \\
 \midrule
\multirow{2}{*}{20\%} & Magnitude Prune & 29.33 & 12.72 & 15.68 & 36.78 & 14.07 \\
 & \method & \textbf{16.57} & \textbf{9.74} & \textbf{10.08} & \textbf{16.78} & \textbf{11.07} \\
  \midrule
\multirow{2}{*}{30\%} & Magnitude Prune &  108.23 & 86.48 & 43.42 & 128.04 & 66.48 \\
 & \method &\textbf{36.20} & \textbf{32.48}& \textbf{20.26} & \textbf{32.15} & \textbf{29.79} \\
  \midrule
\multirow{2}{*}{20\%} & SliceGPT & 11.25  & 6.92 & \textbf{6.76} & 8.11 & 7.67 \\
 & PCA+\method & \textbf{10.24} & 7.01 & 6.98 & \textbf{7.96} & 7.71 \\
  \midrule
\multirow{2}{*}{30\%} & SliceGPT &  19.42 & 9.30 & \textbf{8.69} & 10.73 & 9.14 \\
 & PCA+\method & \textbf{16.46} & 9.26 & 8.89 & 10.68 & 9.20 \\
 \bottomrule\bottomrule \\
\end{tabular}
\label{tab:model-pruning}
\end{table*}

\subsection{Pruning Comparisons}

We apply \method~ atop of two pruning methods. 
The first method is a simple magnitude-based pruner where at each Transformer sublayer, neurons with the lowest average $\ell_2$ norm on calibration data are pruned. 
Neuron indices can differ across each layer by projecting residual connections too, as detailed in section \ref{sec:applying_maps}, as we found this to be superior pruning the same indices throughout the entire model, as proposed in LLM-Pruner as ``channel-pruning'' \cite{ma2023llm}. 
When adding our method to this baseline, Magnitude Pruning, we select the non-pruned neurons as the target support of \method, to get a direct comparison of pruning versus OT-based merging in this setting. This means that to select the neurons that comprise the $d_\text{new}$ support, we keep the neurons with highest average $\ell_2$ norm. 

The second pruning method is SliceGPT, which first applies PCA to reproject features, and then prunes neurons with least total variance \cite{ashkboos2024slicegpt}. This method reflects a popular and competitive approach within width pruning. We directly compare our method to SliceGPT by reprojecting activations by their PCA basis before applying \method~ to reduce neuron width, and call this method PCA+\method. Neurons are selected by $l_1$ norm. We differ from their method by forgoing the slicing matrix and instead using our transport map in its place. We use Sinkhorn regularization $\lambda = 0.1$ for both settings. 

\section{Results}
\label{sec:results}

\subsection{Language Modeling}

Table \ref{tab:model-pruning} presents language modeling results for the uncompressed models, existing pruning techniques, and our method applied atop the pruning methods. 
In applying \method~atop Magnitude Prune, \method~substantially lowers model perplexity by preserving and leveraging signal from neurons that would otherwise be discarded.
While Magnitude Prune only retains the neurons with the highest $\ell_2$ activation norms, \method~ uses these high-norm neurons as a support set and redistributes the full neuron width onto them based on similarity.
Although this redistribution can sometimes reduce the influence of neurons that would have been retained by Magnitude Prune, the advantage of preserving signal from the pruned neurons more than compensates for this effect, as seen by the perplexity reduction.

In comparing the PCA+\method~ results with SliceGPT, we see that for the Llama-3.1 8B model, our method reduces the perplexity by almost 3 points at 30\% width reduction, and over 1 point for 20\% width reduction. For the other models, while Mistral-7B does not improve from using PCA+\method, other models either maintain similar performance or improve with PCA+\method. Given that \method~can improve performance with the difference in objectives to reduce neuron width, we are able to demonstrate an interesting limitation of SliceGPT. 

Using PCA to reduce neuron width in SliceGPT is the $\ell_2$-optimal strategy for linear dimensionality reduction, as it solves the $\ell_2$ reconstruction problem between original and down-projected activations. 
However, by outperforming SliceGPT in a comparable setting, we can demonstrate that minimizing $\ell_2$ activation distance does not always correlate with improved downstream performance. This indicates that relying on $\ell_2$ activation distance as a proxy for model performance, which is common in compression applications \cite{dong2017learning, frantar2022optimal}, may miss other aspects of model behavior. 

\begin{table*}[h]
\caption{Zero shot performance (\% Accuracy) of models on downstream tasks under different width removal settings. Bolded results reflect the higher result on model and sparsity comparisons. \\ }
\centering
\resizebox{\textwidth}{!}{
\begin{tabular}{lllcccccc}
\toprule\toprule
Model  & Sparsity & Method & ARC-C & ARC-E & HellaSWAG & PIQA & Winogrande & Average \\
\midrule 
 \multirow{5}{*}{Llama-3.1-8B} & 0\% & - & 
53.50 & 81.14 & 78.88 & 81.23 & 73.48 & 73.65 \\    \cmidrule{2-9}
& \multirow{2}{*}{20\%} & SliceGPT & 
40.61 & 68.22 & 59.53 & 73.23 & 62.12 & 60.74 \\
& & PCA+\method & 
39.42 & 67.63 & 60.13 & 73.01 & 64.48 & \textbf{60.94} \\   \cmidrule{2-9}
& \multirow{2}{*}{30\%} & SliceGPT & 
34.13 & 58.67 & 45.73 & 67.03 & 53.91 & 51.89 \\
& & PCA+\method & 
34.47 & 57.41 & 47.53 & 67.90 & 55.96 & \textbf{52.65 }\\     \cmidrule{1-9}
\multirow{5}{*}{Llama-3.1-70B} & 0\% & - & 
65.10 & 86.66 & 84.94 & 84.28 & 79.56 & 80.11 \\   \cmidrule{2-9}
& \multirow{2}{*}{20\%} & SliceGPT & 
55.20 & 80.43 & 69.19 & 79.65 & 72.93 & \textbf{71.48}  \\
& & PCA+\method & 
45.39 & 69.28 & 69.28 & 74.59 & 72.22 & 67.43\\   \cmidrule{2-9}
& \multirow{2}{*}{30\%} & SliceGPT & 
50.26 & 73.91 & 63.85 & 77.20 & 70.24 & \textbf{67.09} \\
& & PCA+\method & 
49.91 & 73.61 & 56.19 & 76.28 & 68.98 & 65.00 \\     \midrule

\multirow{5}{*}{Mistral-7B } & 0\% & - & 
52.22 & 78.20 & 80.47 & 82.26 & 73.80 & 73.39 \\    \cmidrule{2-9}
& \multirow{2}{*}{20\%} & SliceGPT & 
42.83 & 69.91 & 61.80 & 75.57 & 64.33 & 62.89 \\
& & PCA+\method & 
42.66 & 69.95 & 62.31 & 75.35 & 65.43 & \textbf{63.14} \\   \cmidrule{2-9}
& \multirow{2}{*}{30\%} & SliceGPT & 
36.18 & 60.69 & 49.04 & 70.02 & 58.88 & \textbf{54.96} \\
& & PCA+\method & 
34.04 & 58.54 & 47.65 & 69.10 & 59.04 & 53.67  \\     \cmidrule{1-9}
\multirow{5}{*}{Mistral-12B } & 0\% & - & 
57.85 & 81.65 & 82.79 & 82.37 & 73.64 & 75.66 \\   \cmidrule{2-9}
& \multirow{2}{*}{20\%} & SliceGPT & 
29.86 & 46.51 & 53.19 & 68.17 & 59.43 & 51.43 \\
& & PCA+\method & 
43.94 & 71.09 & 60.88 & 75.24 & 64.56 & \textbf{63.14} \\   \cmidrule{2-9}
& \multirow{2}{*}{30\%} & SliceGPT & 
26.96 & 40.24 & 42.88 & 63.11 & 56.51 & 45.94 \\
& & PCA+\method & 
33.96 & 57.83 & 46.71 & 69.26 & 60.77 & \textbf{53.71} \\     \midrule

\multirow{5}{*}{Phi-4 14B} & 0\% & - & 
56.06 & 72.77 & 81.90 & 81.34 & 76.72 & 73.76 \\    \cmidrule{2-9}
& \multirow{2}{*}{20\%} & SliceGPT & 
54.61 & 76.09 & 73.08 & 80.30 & 71.11 & 71.04 \\
& & PCA+\method & 
56.23 & 79.29 & 72.65 & 79.76 & 72.45 & \textbf{72.08} \\   \cmidrule{2-9}
 & \multirow{2}{*}{30\%} & SliceGPT & 
49.83 & 74.12 & 64.57 & 76.33 & 69.38 & \textbf{66.84 }\\
 & & PCA+\method & 
48.55 & 75.17 & 63.39 & 76.44 & 68.67 & 66.44  \\   
\bottomrule\bottomrule \\
\end{tabular}
}
\label{table:zero-shot-results}
\end{table*}

 \subsection{Zero-Shot performance on downstream applications}

Given that PCA+\method~outperforms Magnitude Pruning-based \method~in language modeling, we evaluate only SliceGPT and PCA+\method~on zero-shot tasks. Accuracy results across all 5 tasks and an average accuracy are found in Table \ref{table:zero-shot-results}. As seen in the table, several model/sparsity combinations are improved using PCA+\method, including Mistral-Nemo 12B, which sees a substantial 11\% average accuracy improvement across tasks at 20\% sparsity. Additionally, using PCA+\method~improves results for Phi-4 at 20\%, which is remarkably robust to both our compression method and SliceGPT. At 20\% sparsity, this model retains 98\% of its original average zero-shot performance, and even sees higher accuracy on the ARC-Easy and Challenge sets after compression versus in the base model.

We also observe model-specific differences in both language modeling and zero-shot performance, with certain models being very receptive to this signal recombination approach, and others benefiting only slightly or not at all. 
These variations potentially stem from training differences across state-of-the-art LLMs, which can affect their compatibility with merging-based compression. 
To investigate this further, we plot the effective rank of activations (at 99\% variance) for Llama-3.1-8B, enhanced by \method, and Mistral-7B, which follows SliceGPT-like trends. These results, captured during compression, are detailed in Appendix \ref{app:effective_rank}.
We find that Mistral activations have a much lower effective rank than Llama activations in earlier layers, suggesting that Mistral activations contain most of its signal within its first few neurons after PCA reprojection. 
Despite this concentration, we find our method remains sensitive to varying-width reduction strategies across layers, mirroring findings in \citet{ashkboos2024slicegpt}.
In practice, we recommend evaluating both approaches and, where possible, lightly tuning our method, which is relatively inexpensive due to its transform-in-place nature.

\subsection{Analysis}

\paragraph{Compute cost versus performance}

\begin{figure*}[t]
    \centering
    \begin{minipage}[t]{0.32\textwidth}
        \centering
        \includegraphics[width=\linewidth]{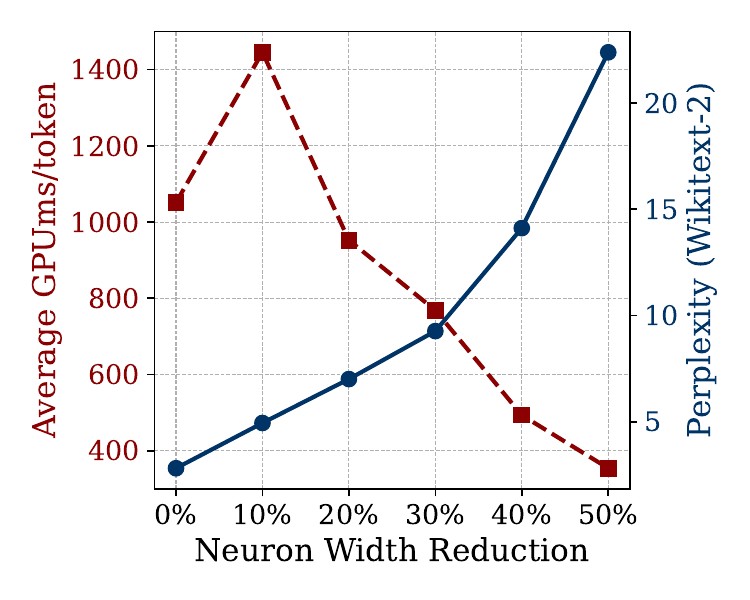}
        \caption{Sparsity vs performance tradeoff across different levels of compression on Llama-3.1-70B. Sparsity (\%) indicates how much width is removed from weight matrices. After 20\%, real world compute cost reduction can be observed.}
        \label{fig:latency}
    \end{minipage}
    \hfill
    \begin{minipage}[t]{0.32\textwidth}
        \centering
        \includegraphics[width=\linewidth]{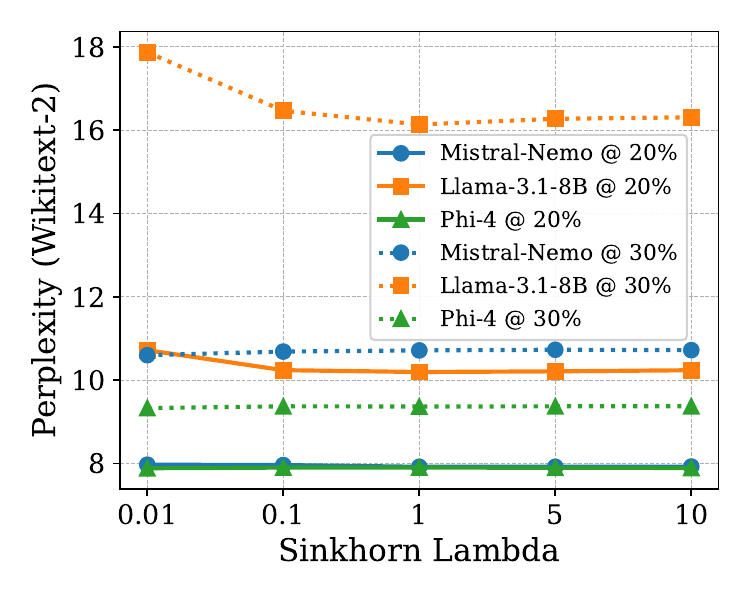}
        \caption{Performance on Wikitext-2 at 20\% and 30\% width reduction with different entropic regularization $\lambda$. Our method is not sensitive to the amount of regularization for these models.}
        \label{fig:ablate_lambda}
    \end{minipage}
    \hfill
    \begin{minipage}[t]{0.32\textwidth}
        \centering
        \includegraphics[width=\linewidth]{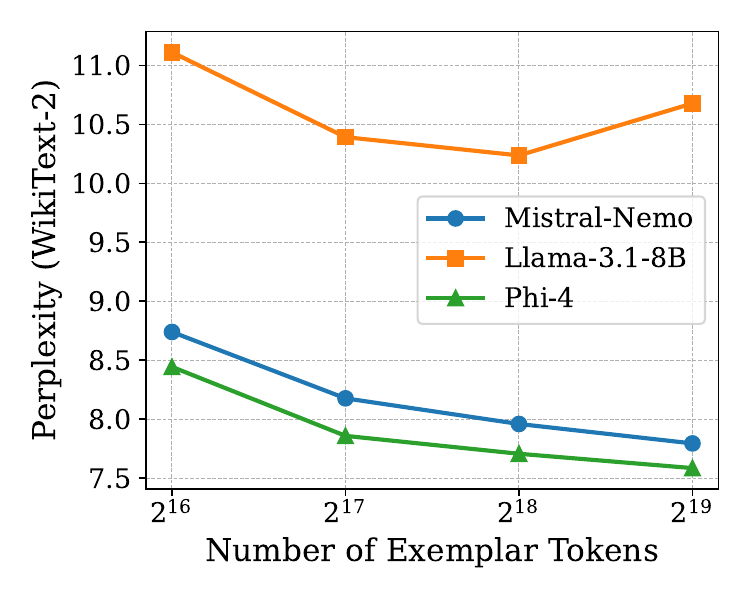}
        \caption{Performance at 20\% sparsity with varying exemplar tokens. After approximately 130K tokens, the returns on using more calibration data appear diminishing.}
        \label{fig:ablate_data_amt}
    \end{minipage}
\end{figure*}

As our method is intended to reduce memory use and improve latency, we measure the total compute used in inference: we find the minimum number of 32GB V100 GPUs the compressed Llama-3.1-70B model can run on without GPU-offloading, and measure the latency in ms/token during inference on sequences of 128 tokens \cite{ashkboos2024slicegpt}. 
This gives us the total compute cost in GPUms/token. 
We compress at different levels of width reduction, and evaluate both perplexity and inference cost, and report this trade-off in in Figure \ref{fig:latency}. After 20\% width reduction, the overall latency begins to improve; the small region of increased cost is incidental to the cost involved in reprojecting the residual stream in both SliceGPT and our method.
\noindent

\paragraph{Sinkhorn regularization parameter}

We select $\lambda = 0.1$ as the regularization parameter in Equation \ref{eq:1}. 
To understand the effect of different values of $\lambda$ on our method, we evaluate three of our models on 5 different $\lambda$ values, ranging from 0.01 to 10.
As seen in Figure  \ref{fig:ablate_lambda}, the amount of regularization does not significantly impact downstream language modeling performance, suggesting that multiple weighting strategies can successfully recombine neurons. 
However, minor variations in $\lambda$ can result in some perplexity differences in the Llama-3.1-8B model as seen at $\lambda=0.01$. 
While optimal $\lambda$ values may vary due to training differences across models, the results presented here indicate a broad range of effective settings.

\paragraph{Calibration Data}

Similar to prior work that examines the behavior of neurons for alignment or pruning \cite{singh2020model, ainsworthgit, ashkboos2024slicegpt}, our method uses a small amount of relevant calibration data (128 samples) to determine the behavior of individual neurons. 
We evaluate the sensitivity of our method to the amount of calibration data used by testing the use of 32 samples, 64 samples, 128 tokens, and 256 samples to compute transportation maps. Results are shown in Figure \ref{fig:ablate_data_amt}.

While \method~ is not notably sensitive to the amount of calibration data used, we recommend using at least 64 samples for computing transport maps. This robustness is a strength, as computing the cost matrices $C$ from Equation \ref{eq:1} can become computationally expensive with large datasets. Additionally, this robustness suggests that a small amount of exemplar data is sufficient to capture meaningful neuron behavior, as reflected in the generalization ability of our compressed models.

\section{Conclusion and Future Work}

In this work, we introduce \method, a novel model compression method that frames neuron width reduction as an optimal transport problem in order to re-project a layer's neurons, specifically avoiding the pruning of any particular neuron. 
In taking this approach, we are able to demonstrate improved performance across several LLMs compared to pruning-based alternatives, demonstrating the importance of incorporating the full signal of a layer and the potential insufficiencies of pruning.
We find some models to be hugely receptive to the addition of a merging-based objective in width reduction, with Phi-4 maintaining almost $98\%$ of zero-shot performance on downstream tasks while improving on some individual tasks at 20\% sparsity, and Mistral-Nemo achieving an over 10\% gain in average accuracy across tasks over SliceGPT. 
We also demonstrate a technical extension of computational invariance in Transformers by demonstrating that invertible matrices can be applied to Transformer weights via a QR-decomposition step, and without changing the model function. 
Several directions merit future exploration. The model-specific variations we observe suggest that intrinsic model properties may influence OT-based compressibility, understanding these factors could enable more adaptive compression strategies. 
More broadly, discrete optimal transport offers a principled framework for parameter transformation that may extend beyond compression.

\newpage

\section*{Impact Statement}

In reducing the memory cost and latency of LLM inference, the broader impacts of our work are several.
Firstly, reducing the latency and power needed to run these models reduces both the monetary and environmental costs of LLMs.
These downstream effects are intended positive consequences resulting from our work. 

Additionally, reducing the computational resources needed to run LLMs can have both positive and negative consequences with respect to accessibility. Much like LLMs may have positive uses in society, they have many negative uses as well. In improving the accessibility of these models, we contribute to amplifying this effect, in reaching both good- and poor-intentioned users.

\bibliography{icml}
\bibliographystyle{icml2026}

\newpage
\appendix
\onecolumn

\section{Simple explanation of QR Decomposition invariance}

\label{app:qr_proof}
From Section \ref{method:qr}, we provide an equality statement in Equation \ref{eq:qr_invariance}. We include a short and simple proof for completeness. As a reminder, for model dimension $d_{\text{orig}}$ we have input vector $\mathbf{x} \in \mathbb{R}^{d_\text{orig}}$, $Q$ denotes orthogonal matrices where $Q^TQ = QQ^T = I$, $T = QR$ via QR decomposition, and $T$ is invertible.
\begin{align} 
\text{RMSNorm}(\mathbf{x}Q) RT^{-1} &= \frac{\mathbf{x}Q}{\lVert \mathbf{x}Q \rVert} RT^{-1} && \text{via RMSNorm definition} \\
&=\frac{\mathbf{x}QRT^{-1}}{\lVert \mathbf{x} \rVert}  && \text{via $\lVert xQ \rVert = \sqrt{xQ^TQx} = \lVert x \rVert $} \\
&= \text{RMSNorm}(\mathbf{x}) && \text{via $QR = T$ substitution}
\end{align}

\section{QR Decomposition invariance for rectangular maps}
\label{app:rectangular_qr_proof}

We will prove a theorem related to the RMSNorm, and then discuss what the result dictates in terms of invariance to RMSNorm in the rectangular transport map $T$ case. 

\begin{theorem}[QR invariance for rectangular maps]
Let $T \in \mathbb{R}^{d_{\text{orig}} \times d_{\text{new}}}$ be QR-decomposed into
$T = QR$, where semi-orthogonal $Q \in \mathbb{R}^{d_{\text{orig}} \times d_{\text{new}}}$ has orthonormal columns
($Q^T Q = I_{d_{\text{new}}}$) and $R \in \mathbb{R}^{d_{\text{new}} \times d_{\text{new}}}$ is upper
triangular and invertible. Then,
\begin{align}
\text{RMSNorm}(xQ)RT^{\dagger}
= \frac{xQQ^T}{\|xQQ^T\|}.
\end{align}
\end{theorem}

\begin{proof}
We first note that $Q$ is semi-orthogonal, therefore
\begin{align}
\|xQ\|
&= \sqrt{xQQ^T x^T}
= \sqrt{xQ(Q^TQ)Q^T x^T}
= \|xQQ^T\|.
\end{align}

Next, using the QR decomposition of $T$, the Moore--Penrose pseudoinverse is
\begin{align*}
T^{\dagger}
&= (T^T T)^{-1} T^T  \\
&= (R^T Q^T Q R)^{-1} R^T Q^T  \\
&= (R^T R)^{-1} R^T Q^T \\
&= R^{-1} Q^T.
\end{align*}

Substituting into the expression of interest gives
\begin{align}
\text{RMSNorm}(xQ)RT^{\dagger}
&= \frac{xQ}{\|xQ\|} R (R^{-1} Q^T) \\
&= \frac{xQ Q^T}{\|xQ\|} \\
&= \frac{xQ Q^T}{\|xQ Q^T\|},
\end{align}
which completes the proof.
\end{proof}

Given this result, we see that the RMSNorm is invariant to our constructed maps up to orthogonal projection onto the column space of $Q$.

\section{Layer-wise analysis of effective rank}

\begin{figure}[h]
    \centering
    \includegraphics[width=0.5\linewidth]{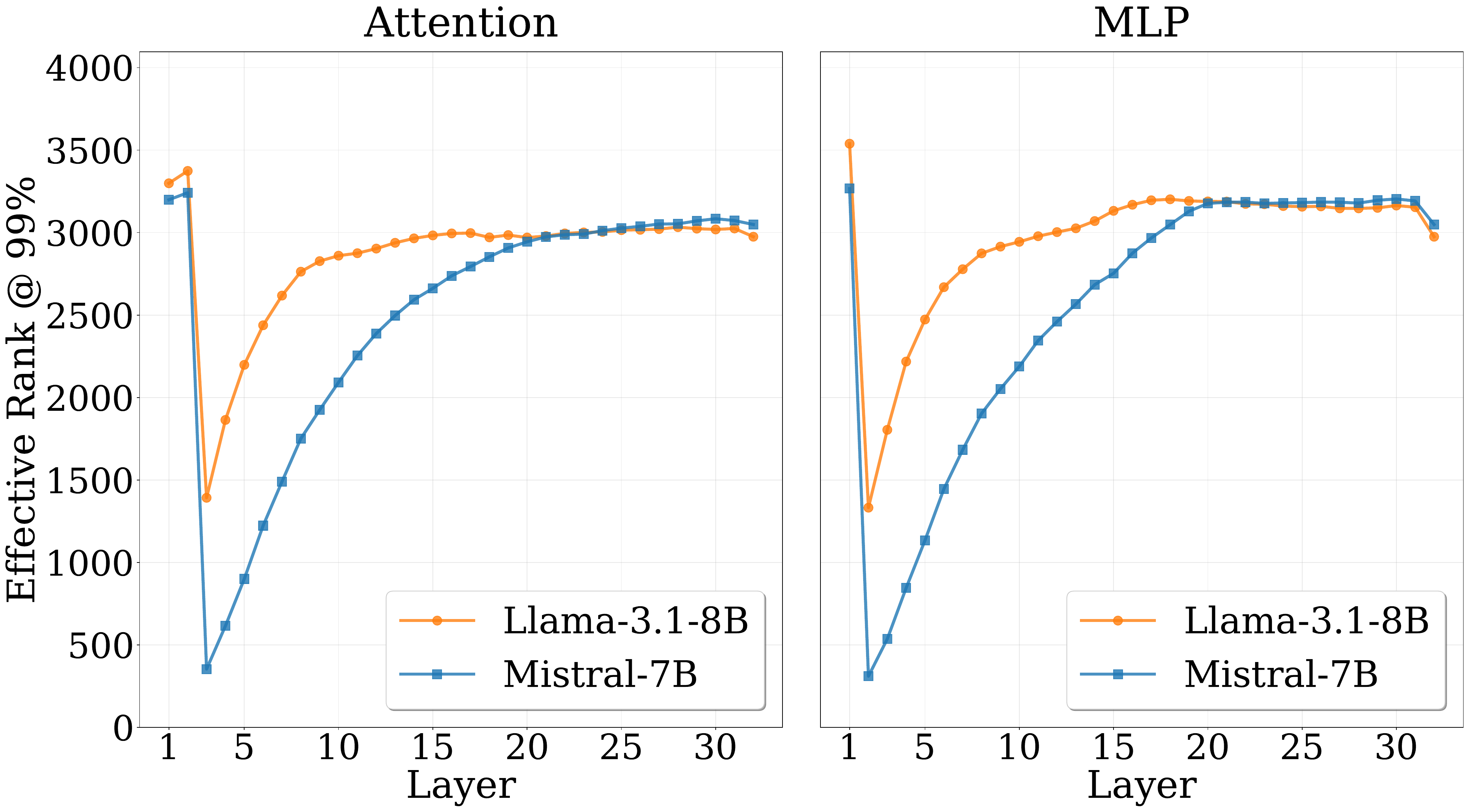}
    \caption{Effective rank at 99\% variance of Llama-3.1-8B and Mistral-7B activations at each layer during PCA+\method~application for 20\% width reduction. We report values for both the output of attention and the MLP, after the residual connection. We find that Mistral reaches a much smaller effective rank in earlier layers, suggesting that it contains less distributed signal across neurons after PCA reprojection.}
    \label{fig:effective_rank}
\end{figure}

For a data matrix of $n$ items of dimension $d_\text{orig}$, we compute effective rank as the following, assuming we have ordered eigenvalues $\{ \sigma_i \}_{i\in [1,d_\text{orig}]}$, and threhold $0 < \tau < 1$: 
$$k = \arg\min_{k'} \frac{\sum_{i=1}^{k'} \sigma_i }{\sum_{i=1}^{d_\text{orig}} \sigma_i} > \tau$$
In other words, we find the minimum index $k$ such that the cumulative sum of the first $k$ eigenvalues exceeds ratio $\tau$. In this analysis, we fix $\tau = 0.99$. 
We compute the effective rank for two models, Llama-3.1-8B and Mistral-7B across all 32 layers, at both the output of the attention sublayer and the output of the MLP sublayer. 
Both models have true model dimension of 4096. 
We report our results in Figure \ref{fig:effective_rank}, and find that Mistral has significantly reduced effective rank for many early layers as compared to Llama, suggesting that Llama has more disperse signal across its PCA-reprojected neurons. 

\label{app:effective_rank}

\end{document}